\documentclass[a4paper,times]{article}
\usepackage{blindtext}

\usepackage[english]{babel}

\usepackage{amsmath}
\usepackage{graphicx}
\usepackage{authblk}

\usepackage[margin=1in]{geometry}

\title{Control of a simulated MRI scanner with deep reinforcement learning}

\begin{document}

\author[1]{Simon Walker-Samuel}
\affil[1]{Centre for Computational Medicine, University College London, London, UK}

\maketitle

\begin{abstract}
Magnetic resonance imaging (MRI) is a highly versatile and widely used clinical imaging tool. The content of MRI images is controlled by an acquisition sequence, which coordinates the timing and magnitude of the scanner hardware activations, which shape and coordinate the magnetisation within the body, allowing a coherent signal to be produced. The use of deep reinforcement learning (DRL) to control this process, and determine new and efficient acquisition strategies in MRI, has not been explored. Here, we take a first step into this area, by using DRL to control a virtual MRI scanner, and framing the problem as a game that aims to efficiently reconstruct the shape of an imaging phantom using partially reconstructed magnitude images. Our findings demonstrate that DRL successfully completed two key tasks: inducing the virtual MRI scanner to generate useful signals and interpreting those signals to determine the phantom's shape. This proof-of-concept study highlights the potential of DRL in autonomous MRI data acquisition, shedding light on the suitability of DRL for complex tasks, with limited supervision, and without the need to provide human-readable outputs.
\end{abstract}



\section*{Introduction}

Magnetic resonance imaging (MRI) is widely used in clinical medicine, with scanners located in most general hospitals. Patients are positioned inside a large magnetic field to exploit the interaction between hydrogen nuclei (mainly water) and radiofrequency pulses to create a measurable radiofrquency signal. By encoding the position of this signal according to its frequency and phase, images can be formed showing the inside of the human body. The characteristics of those images, such as contrast, signal-to-noise, resolution, etc. are determined by the acquisition (or pulse) sequence, which describes the timing of repeated hardware activations, including radiofrequency pulses and static field gradients. The two main categories of acquisition sequence are the spin echo and gradient echo, but these also include numerous subcategories.

The broad range of acquisition sequences now available as standard on clinical MRI scanners has been developed in response to clinical need, such as the detection or diagnosis of particular cancers. They are optimised to provide contrast between tissues of interest with acceptable signal-to-noise characteristics to make the images interpretable to human radiologists, or to quantify tissue properties such as blood flow or water diffusion. 

Acquisition sequences are designed with human interpretability formost, although this requirement has been questioned with the increasing prominence of deep learning. Automated image interpretation is widely predicted to transform the practice of radiology, and most approaches use supervised learning, in which deep neural networks are trained to categorise or detect features within standard clinical images. However, unsupervised or semi-supervised deep learning approaches also exist, such as reinforcement learning, which have not been widely explored in the context of MRI data acquisition.

In deep reinforcement learning (DRL), multi-layered artificial neural networks are used to provide highly adaptive, efficient, and generalizable architecture for learning representations of data with multiple levels of abstraction. DRL is a semi-supervised algorithm that uses artificial neural networks to iteratively guide the actions taken within an environment to maximise a notional reward. 

Given that the physics underpinning MRI acquisitions is well-defined, this provides an environment for control with DRL. We aimed here to use DRL to control a virtual MRI scanner. The physics underpinning the interaction between tissue water and MRI acquisition sequences is well-described by the Bloch equations, allowing straightforward simulation of a virtual MRI scanner. Control of our virtual scanner was cast in the form of a game in which the aim was to guess the shape of an object (an imaging phantom), as quickly as possible, guided by partially reconstructed magnitude images. The aim of this study was therefore to determine if we could use a DDPG algorithm to determine a solution to this relatively simple MRI problem, with no external supervision. This requires the DDPG algorithm to perform two separate tasks: 1) learn how to induce the virtual MRI scanner to generate useful signals and 2) learn how to interpret those signals and determine the shape of the phantom. Whilst ostensibly simple, this problem requires multiple levels of abstraction, which we hypothesised deep reinforcement learning would be well suited to.

\textit{This study was originally presented at the 2019 meeting of the International Society for Magnetic Resonance in Medicine (ISMRM).} [1]

\section*{Methods}

\subsection*{BlochGame}
The reinforcement learning environment was cast in the form of a game named BlochGame (Figure 1). In each round, a square or circular virtual phantom was generated in the centre of a 32$\times$32 matrix. Phantom sizes ranged from 2 to 10 pixels, with $T_1$/$T_2$ = 1300/20 ms. Background areas had $T_1$/$T_2$ = 3000/2000 ms to simulated an aqueous environment. TR, TE, flip angle and phase-encoding step could be controlled, for each k-space line acquisition in a gradient echo-type sequence. After each line acquisition, the DRL agent could guess the shape of the phantom, guided by the partially-reconstructed image. For a correct shape guess ($G_+$), no guess ($G_0$), and incorrect guess ($G_-$), the following reward scheme was adopted:
\begin{equation}
\begin{aligned}
G_+ &= A e^{-Rt/t_{max}} \\
G_0 &= 1 \\
G_- &= B
\end{aligned}
\end{equation}

where $t_{max}$ is the maximum time for each game (set to 5 minutes), after which the game was abandoned with a score of zero. R is the rate of score decay with time $t$ (set to 1 s-1), $A$ is the correct guess reward weighting (set to 100) and $B$ is the incorrect guess reward weighting (set to -10). The aim of the game was to acquire data efficiently and sparsely, whilst maintaining accurate shape guesses.

This game design was chosen as we wanted to determine strategies that enabled data to be acquired rapidly, whilst maintaining sufficient contrast/signal-to-noise characteristics to enable the shape of the phantom to be determined. For example, for the $T_1$ and $T_2$ we assigned to the phantom and background, a long TE and TR would be favourable to generate image contrast but would result in a longer acquisition time. Moreover, as spoiling is not included in the acquisition sequence, the influence of residual magnetisation must also be taken into consideration, again favouring a longer TR.

\begin{figure*}
\centering
\includegraphics[width=\linewidth]{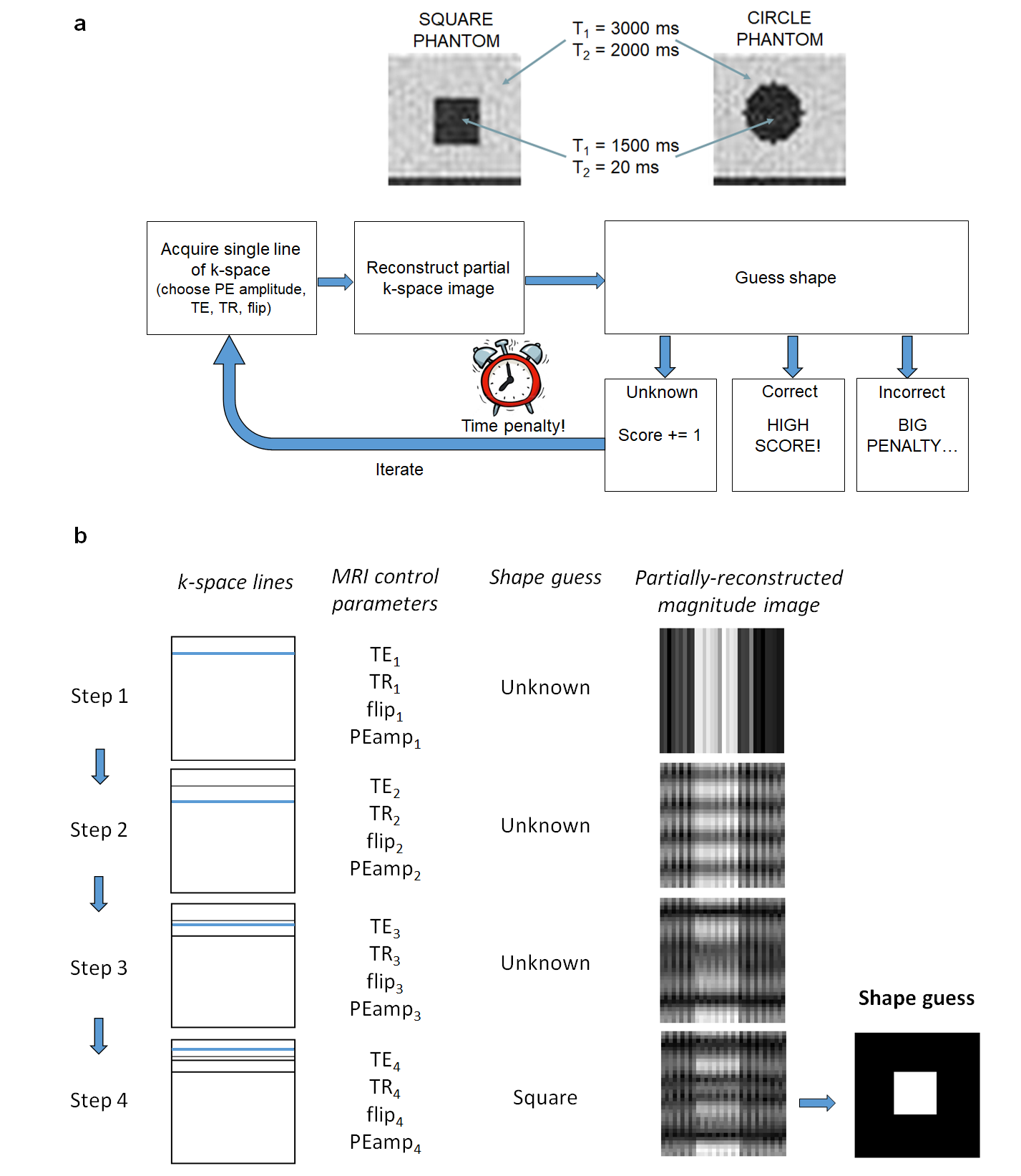}
\caption{\label{fig:fig1}a) A schematic overview of BlochGame, a virtual environment for trialling deep reinforcement learning for MRI scanner control. The aim  of BlochGame is to acquire MRI data to determine which test object (phantom) is in the scanner – either a square or a circle – as quickly as possible. b) An example BlochGame sequence, where at each step, the values of parameters that control the MRI scanner acquisition are chosen. This includes the phase encoding gradient amplitude, which controls the k-space line location to be filled. After each k-space line acquisition, k-space is inverse Fourier transformed and a magnitude image produced, from which the shape of the phantom can be guessed.}
\end{figure*}

\subsection*{MRI signal simulation}
MRI data were generated using the following solutions to the standard Bloch equations. In the rotating reference frame these are given by
\begin{equation}
\frac{d\vec{M}}{dt} = \gamma \vec{M} \times \vec{B_e} - \begin{pmatrix} M_x/T_2 \\ M_y/T_2 \\ (M_z-M_0)/T_2 \end{pmatrix}
\end{equation}
where $\gamma$ and $\vec{B_e}$ are the gyromagnetic ratio and effective magnetic field in the rotating frame of reference, respectively. 
\begin{equation}
\begin{split}
B_e(x,y,z,t) = & \Delta B_z(x,y,z)  + (G_x(t) x + G_y(t) y + G_z(t)z)\cdot\hat{z} + B_{1x}(t)\cdot\hat{x} + B_{1y}(t)\cdot\hat{y}
\end{split}
\end{equation}
where $\delta B_z(x,y,z)$ is the $z$ component of the inhomogeneous magnetic field, $G_x(t)$, $G_y(t)$, and $G_z(t)$ are magnetic field gradients, and $B_{1x}(t)$ and $B_{1y}(t)$ are the two orthogonal components of the RF magnetic field in the rotating frame of reference. RF excitations were modelled as instantaneous rotations, followed by 2 ms of free precession.

A Bloch simulator was written in Python 3.5, using a matrix formulation of the Bloch equations, to allow efficient calculation. A simple, virtual gradient echo sequence was constructed for acquisition, constrained to a field of view of 20 $\times$ 20 cm. A phase encoding table was pre-calculated, containing 64 steps, with amplitudes ranging from –3.8 to 3.8 mT/m. Readout gradient amplitude was 0.75 mT/m, with a duration of 5 ms. 

\subsection*{Deep reinforcement learning}
A deep deterministic policy gradient (DDPG) algorithm with actor-critic architecture [2] was implemented in Tensorflow (Python 3.5), with simulated magnitude pixel data input to the network (see Figure 2). The DRL algorithm used an actor-critic network structure, enabling continuous output parameters. The actor network contained six fully-connected convolutional layers (widths 32, 32, 64, 64, 128, 128) and three fully-connected layers, each with ReLU activations. The output layer had five nodes corresponding to PE index, TE, TR, flip angle and shape guess, which were scaled to the range PE index = (0, 31), TE = (5, 20) ms, TR = (25, 2000) ms, flip = (1, 90)° and shape guess = (0, 2). Shape guess and PE index were converted to integer values, with 0, 1 and 2 corresponding to ‘unknown’, ‘circle’ and ‘square’, respectively.

The critic network had a fully-connected state (pixel) input layer containing 1024 nodes, followed by two fully-connected layers with 24 and 48 nodes, respectively. This was mirrored in the action input (5 nodes). Both inputs were combined in a merging layer, followed by a layer containing 24 nodes. The output layer contained a single node. The network was trained for $6 \times 10^6$ games using two Nvidia 1060 GPUs. Training incorporated incorporated experience replay (buffer size of 1000) and an epsilon-greedy exploration policy (9). The resulting network was evaluated for 10,000 games.

\begin{figure}
\centering
\includegraphics[width=\linewidth]{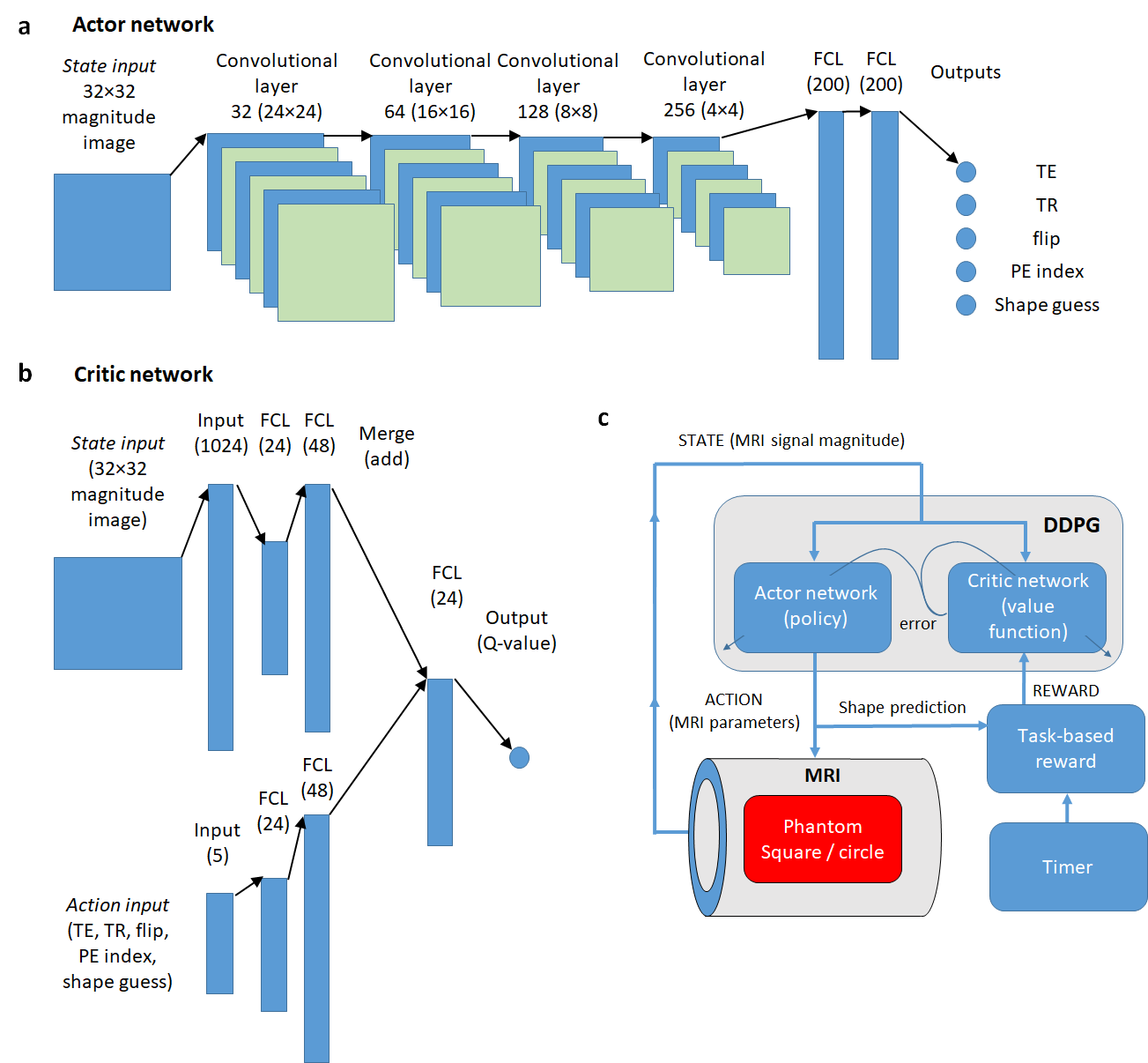}
\caption{\label{fig:fig2}a) Schematic diagram of the deep reinforcement learning algorithm used to control a virtual MRI scanner and determine the shape of a phantom, showing a) the actor network, b) the critic network, and c) an overview of the whole system.}
\end{figure}

\begin{figure}
\centering
\includegraphics[width=\linewidth]{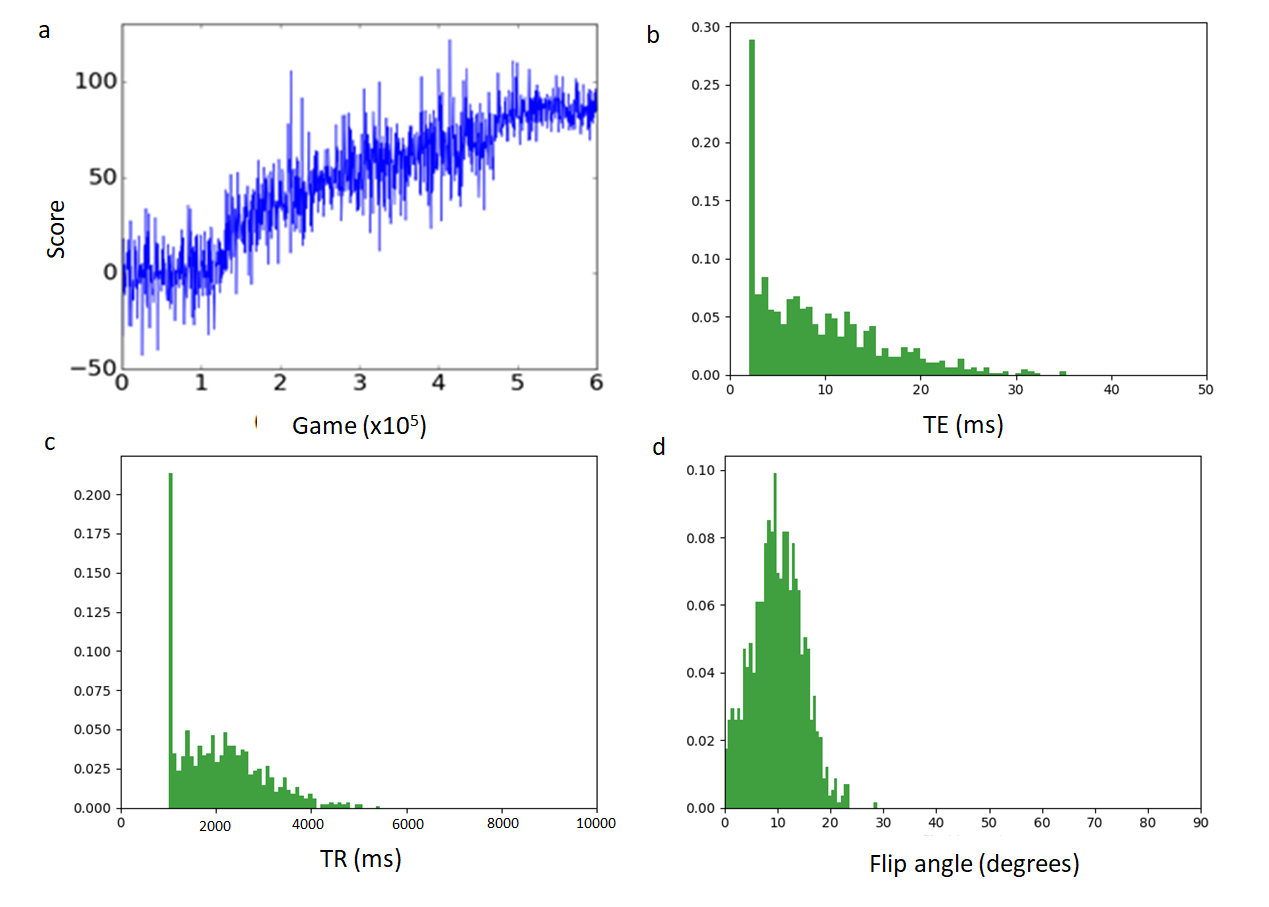}
\caption{\label{fig:fig3}a) DDPG score during training, showing the DDPG algorithm learning improved acquisition strategies with increasing number of games played. b-d) Histograms of b) echo time (TE), c) repetition time (TR) and d) flip angle during 10,000 evaluation games.}
\end{figure}

\section*{Results}
During training, the reward increased from 0 to a maximum of 90, indicating learning by the network (Figure 3a). Following training, square and circular phantoms were distinguished with 99.8\% accuracy. Histograms of each acquisition parameter during the evaluation stage are shown in Figure 3b-c. The main acquisition strategy adopted by the DDPG algorithm was to acquire, on average, 4 outer lines of k-space, with the shortest permitted echo time (5 ms) most often selected, and a mean flip angle of 10 ± 6°. The mean total time taken was 50 $\pm$ 7 ms. Histograms in Figure 3 show the acquisition parameters chosen by BlochNet during 10,000 evaluation games and Figures 4 and 5 show example games played by the trained system. 

Across all 10,000 games, the TR and TE selected provided $T_1$-weighted contrast between the phantom and its aqueous background of between 20-50\%, and with sufficient signal-to-noise characteristics to form an image. This acquisition strategy, learnt independently by the DDPG algorithm, offered a good compromise between acquisition time and signal-to-noise ratio (SNR), allowing it to accurately determine the shape of the phantom; a shorter TR or longer TE would have resulted in lower SNR.

\begin{figure}
\centering
\includegraphics[width=\linewidth]{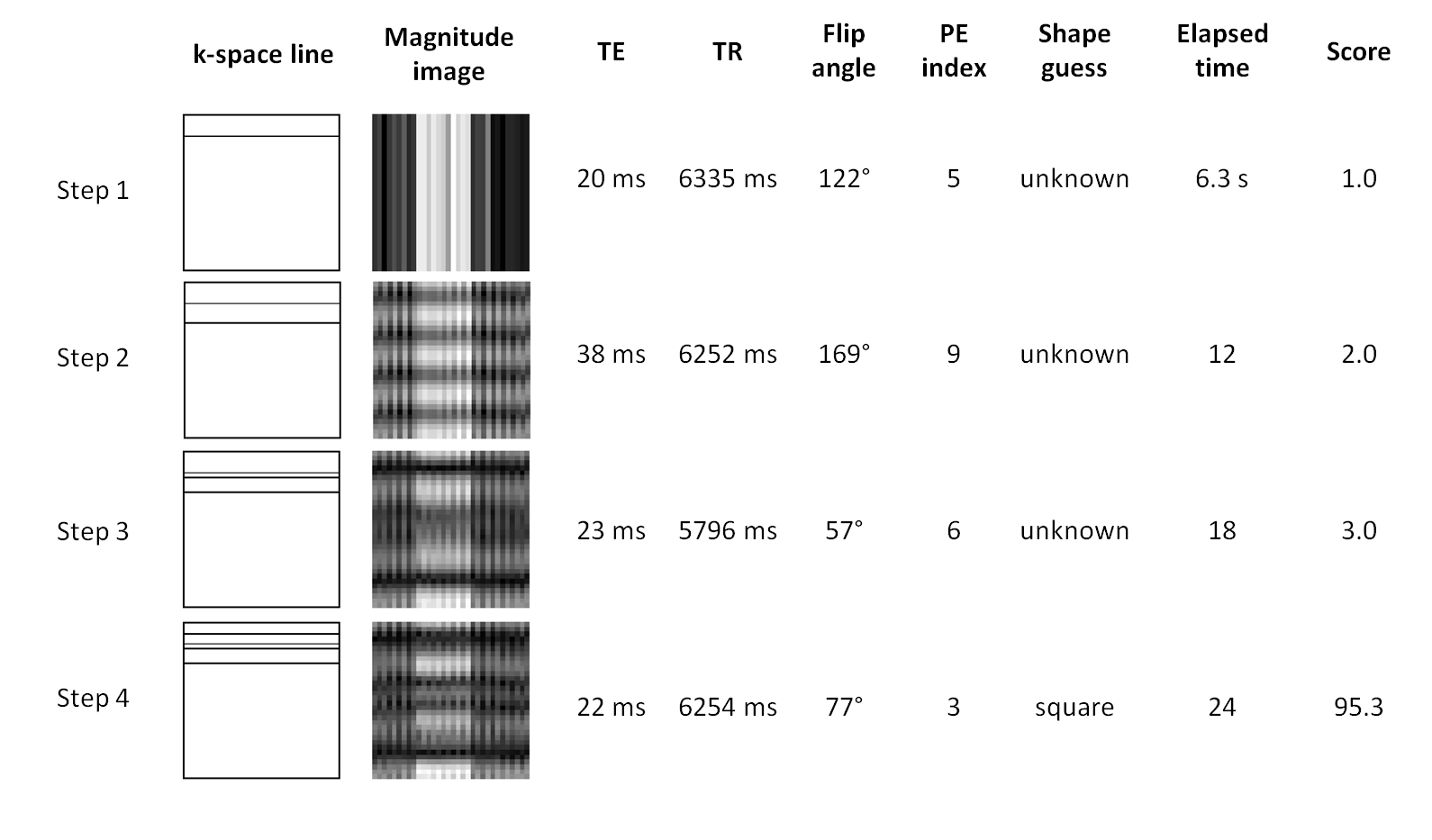}
\caption{\label{fig:fig4}An example sequence from BlochGame, recorded during evaluation, demonstrating the ‘edge detection’ strategy, in which only outer lines of k-space were acquired, thereby acting as an edge detector. Total sequence run time (top to bottom) was 24s, with a score of 134.1.}
\end{figure}

\begin{figure}
\centering
\includegraphics[width=\linewidth]{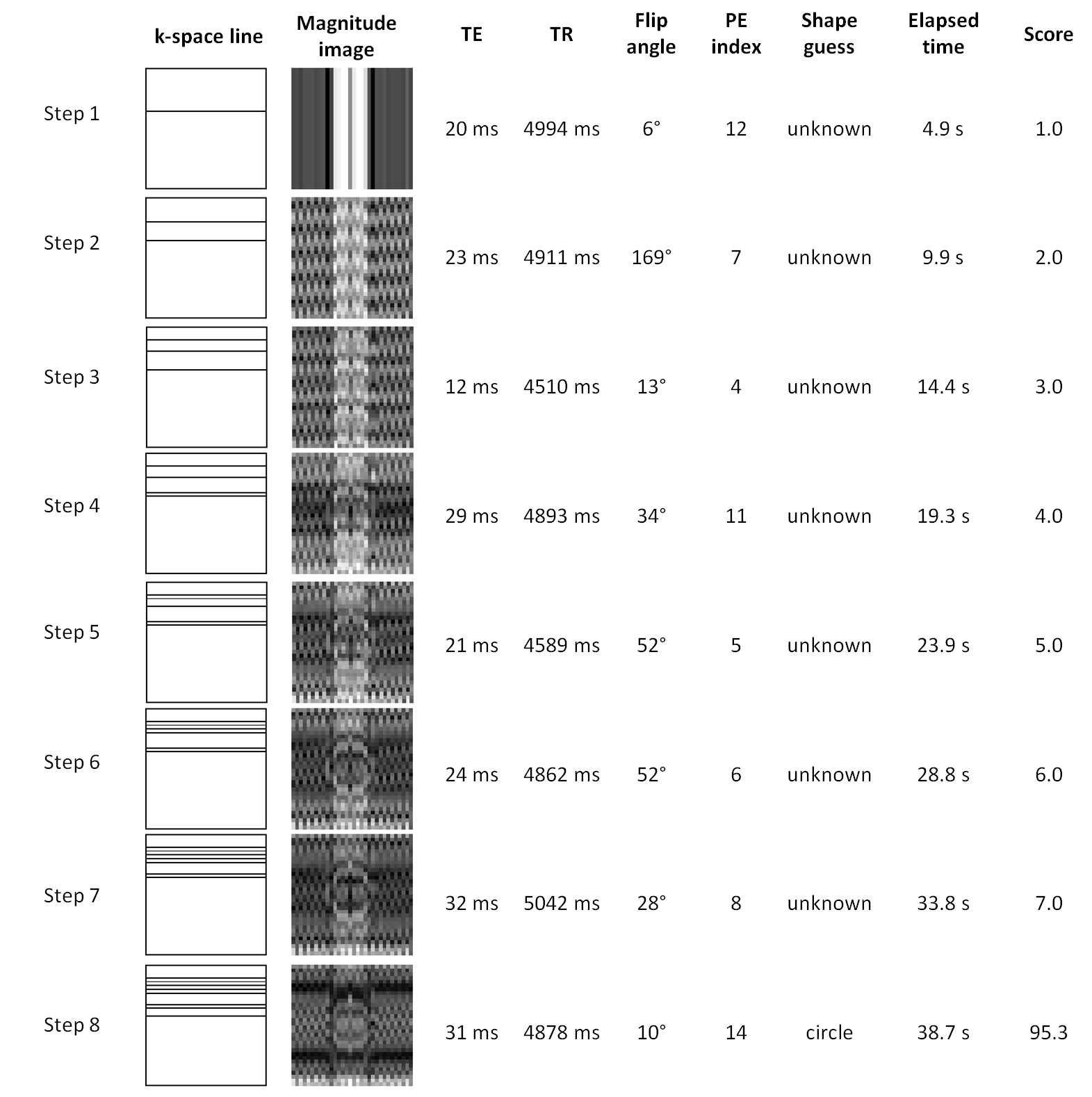}
\caption{\label{fig:fig5}An example sequence from BlochGame, in which a more complex strategy with highly variable flip angle was used. Total sequence run time (top to bottom) was 38.7s, with a score of 82.0.}
\end{figure}

\section*{Discussion}
A key milestone in the rise to prominence of deep learning was DeepMind’s use of deep reinforcement learning to play Atari computer games to human (and even superhuman) levels, using only pixel data as input [3]. They have subsequently gone on to develop algorithms that outperformed human experts in the notoriously complex board game, Go [4, 5] and the computer game StarCraft II [6]. Both games are notoriously complex, and present significant challenges for automated play using conventional machine learning strategies. Subsequently, they developed AlphaFold to predict protein structures [7].

Reinforcement learning has been applied to MRI in several settings, most notably to identify new k-space sampling strategies [8], to identify breast lesions [9] and designing shaped RF pulses [10]. 

We have shown here the feasibility of using deep reinforcement learning to control a virtual MRI scanner and demonstrated that the deep deterministic policy gradient algorithm was able to develop strategies to acquire signal from the scanner with sufficient contrast and signal-to-noise characteristics to differentiate a square from a circular phantom. This was a seemingly simple problem but required the algorithm to navigate multiple linked tasks to achieve the required outcome. Its solution was to acquire data with a short TE and long TR, in a time-efficient manner by only acquiring data from the outer edges of k-space, thereby essentially acting as an edge detector, allowing very high shape guessing accuracy (99.8\%). The form of this solution will have been in part shaped by the rules defined within BlochGame, which were intentionally designed to reward fast acquisition times and accurate shape guessing.

This simple experiment was far removed from the much greater complexity of a real-world scanner. BlochGame featured only four controls (TE, TR, flip angle and phase encoding gradient index), whereas acquisition sequences on modern MRI scanners feature many more. However, it demonstrates proof of principle, and also showed how reinforcement learning can be used to create novel approaches to MRI data acquisition. For example, an interesting observation is the extent to which parameters varied between k-space acquisitions. This is unlike almost all real-world sequences, in which values are usually given a fixed value while k-space is sequentially filled by stepping the phase encoding gradient, giving uniform contrast across the whole image. Instead, DDPG control assigned each k-space line a different set of TE, TR and flip angle, either intentionally or otherwise, allowing mixed-weighting images to be created. Additionally, the DDPG setup also allows early termination of the acquisition once the content of the acquisition is determined, and potentially, in a manner that is not designed to be interpretable by a human.

This is particularly important because determining the optimal method to acquire MRI data,  to address specific clinical questions, is a highly complex, non-linear problem, which can be highly challenging to solve via conventional means. Competing considerations include ensuring reliable signal quality, minimising scan time, and enabling human interpretability. Coupled with the large set of factors that can contribute to the MRI signal, and their complex interactions with tissue physiology, this creates a vast parameter space to be explored. As a result, the development of imaging biomarkers requires substantial resource [11] and, for practical reasons, studies tend to focus on small subsets of this space. 

Each of these challenges can act as a significant barrier to the adoption of new forms of MRI measurements in the clinic, due to: 1) the limited time available to acquire rich, quantitative MRI data, particularly in unwell patients, and in busy clinical environments; 2) difficulty in determining the ‘best’ combination of MRI measurements to provide unequivocal and reproducible diagnostic decisions; 3) difficulties in standardising measurements between institutions; and 4) a lack of data-driven methods and tools to confidently and reproducibly couple MRI measurements to disease states, diagnosis and prognosis. These challenges have meant that the significant potential of many next-generation imaging biomarkers, that could substantially improve patient care, have not yet been realised. 

For reinforcement learning to address these types of challenge, much greater complexity would need to be added into future iterations, including biological mimics or virtual twins in place of imaging phantoms, a broader range of scanner controls, and use three-dimensional data in place of the two-dimensional acquisition used here. Equally, another step would be to run a real-world MRI scanner using a DDPG controller. If successful, a new paradigm in MRI data acquisition could be started, in which scanners operate autonomously, without the need for human pulse programming or data interpretation, and in which training could be undertaken within an additional part of the standard clinical protocol, and fleets of scanners share experience, much like as has been explored in the development of autonomous vehicles.



\section*{References}
\begin{enumerate}
\item Walker-Samuel S, Using deep reinforcement learning to actively, adaptively and autonomously control a simulated MRI scanner. 2019; International Society for Magnetic Resonance in Medicine (Montreal)
\item Lillicrap TP et al. Continuous control with deep reinforcement learning. ArXiv e-prints [Internet]. September 1, 2015; 1509. Available from: http://adsabs.harvard.edu/abs/2015arXiv150902971L.
\item Mnih V, Kavukcuoglu K, Silver D, Rusu AA, Veness J, Bellemare MG, et al. Human-level control through deep reinforcement learning. Nature. 2015; 518:529.
\item Silver D, Huang A, Maddison CJ, Guez A, Sifre L, van den Driessche G, et al. Mastering the game of Go with deep neural networks and tree search. Nature. 2016; 529(7587):484-9.
\item Silver D, Schrittwieser J, Simonyan K, Antonoglou I, Huang A, Guez A, et al. Mastering the game of Go without human knowledge. Nature. 2017; 550(7676):354-9.
\item Vinyals O, Babuschkin I, Czarnecki WM, Mathieu M, Dudzik A, Chung J, et al. Grandmaster level in StarCraft II using multi-agent reinforcement learning. Nature. 2019; 575(7782):350-4.
\item Jumper J, Evans R, Pritzel A, Green T, Figurnov M, Ronneberger O, Tunyasuvunakool K, Bates R, Žídek A, Potapenko A, Bridgland A, Meyer C, Kohl SAA, Ballard AJ, Cowie A, Romera-Paredes B, Nikolov S, Jain R, Adler J, Back T, Petersen S, Reiman D, Clancy E, Zielinski M, Hassabis D. Highly accurate protein structure prediction with AlphaFold. Nature. 2021; 596:583–589.
\item Pineda L, Basu S, Romero A, Calandra R, Drozdzal M. Active MR k-space Sampling with Reinforcement Learning. Lecture Notes in Computer Science book series. 29 September 2020 (LNIP,volume 12262)
\item Deep Reinforcement Learning for Active Breast Lesion Detection from DCE-MRI. Maicas G, Carneiro G, Bradley AP, Nascimento JC, Reid I, Lecture Notes in Computer Science book series. 04 September 2017 (LNIP,volume 10435)
\item Deep reinforcement learning-designed radiofrequency waveform in MRI
Shin D, Kim Y, Oh C, An H, Park J, Kim J, Lee J. Nature Machine Intelligence 2021: 3;985–994.
\item O'Connor J, Aboagye E, Adams J, et al. Imaging biomarker roadmap for cancer studies. Nat Rev Clin Oncol. 2017; 14:169–186
\end{enumerate}
\end{document}